%% file: FaithIR_arXiv.tex
\documentclass[letterpaper]{article} % DO NOT CHANGE THIS
\usepackage[preprint]{aaai2027}  % DO NOT CHANGE THIS
\usepackage[hyphens]{url}  % DO NOT CHANGE THIS
\usepackage{graphicx} % DO NOT CHANGE THIS
\usepackage{natbib}  % DO NOT CHANGE THIS AND DO NOT ADD ANY OPTIONS TO IT
\usepackage{caption} % DO NOT CHANGE THIS AND DO NOT ADD ANY OPTIONS TO IT
\usepackage{algorithm}
\usepackage{algorithmic}
\usepackage{multirow}
\usepackage{amsfonts}

\usepackage{newfloat}
\usepackage{listings}
\DeclareCaptionStyle{ruled}{labelfont=normalfont,labelsep=colon,strut=off} % DO NOT CHANGE THIS
\floatstyle{ruled}
\newfloat{listing}{tb}{lst}{}
\floatname{listing}{Listing}

\usepackage{booktabs}

\title{FaithIR: Rethinking Infrared Image Super-Resolution  from \\ Perceptual Sharpness to Task Relevant Fidelity}
\author{
    Axi Niu\textsuperscript{\rm 1},
    Zhenguo Wu\textsuperscript{\rm 1}\corresponding,
    Kang Zhang\textsuperscript{\rm 2},
    Qingsen Yan\textsuperscript{\rm 1},
    Jinqiu Sun\textsuperscript{\rm 3},
    Yanning Zhang\textsuperscript{\rm 1}
}

\affiliations{
    \textsuperscript{\rm 1} School of Computer Science, Northwestern Polytechnical University, Xi'an, Shaanxi, China\\
    \textsuperscript{\rm 2} School of Electrical Engineering, KAIST, Daejeon, Republic of Korea\\
    \textsuperscript{\rm 3} School of Aeronautics and Astronautics, Northwestern Polytechnical University, Xi'an, Shaanxi, China\\
    wuzhenguo@mail.nwpu.edu.cn
}

\begin{document}

\maketitle

\begin{abstract}

Infrared image super-resolution (IISR) is important for downstream tasks such as object detection and semantic segmentation. Existing IISR methods often produce artificial textures, over-sharpened edges, and spurious high-frequency details that distort authentic thermal structures and semantic information. To address this issue, we propose FaithIR, a faithful infrared super-resolution framework for reliable machine perception. FaithIR consists of a patch-level conditioning branch that captures global thermal and structural information and a pixel-level restoration branch that performs dense local reconstruction under structural guidance. The entire restoration process is performed directly in the pixel domain to preserve infrared-specific structures and task-relevant information. Extensive experiments on FLIR-IISR, M3FD, and FMB demonstrate strong reconstruction fidelity, cross-dataset generalization, and superior performance in object detection and semantic segmentation. These results show that demonstrate that preserving faithful infrared structure preservations is more important for reliable machine perception than merely pursuing perceptual sharpness alone.

\end{abstract}
% Uncomment the following to link to your code, datasets, an extended version or similar.
% You must keep this block between (not within) the abstract and the main body of the paper.
% Make sure that you do not de-anonymize yourself with these links.
% \begin{links}
%     \link{Code}{https://aaai.org/example/code}
%     \link{Datasets}{https://aaai.org/example/datasets}
%     \link{Extended version}{https://aaai.org/example/extended-version}
% \end{links}

% You can remove the copyright notice and ensure that your names aren't shown by including \texttt{submission} option when loading the \texttt{aaai2027} package:

% \begin{quote}\begin{scriptsize}\begin{verbatim}
% \documentclass[letterpaper]{article}
% \usepackage[submission]{aaai2027}
% \end{verbatim}\end{scriptsize}\end{quote}

% The remainder of this document contains the original camera-ready instructions. Any contradiction of the above points ought to be ignored while preparing anonymous submissions.

\section{Introduction}

Infrared image super-resolution (IISR) aims to recover high-resolution (HR) infrared images from low-resolution (LR) observations and plays an important role in downstream tasks such as object detection and semantic segmentation~\cite{huang2025iisrsurvey,fang2024scinet,li2025difiisr}. 
Unlike natural visible-light images, infrared images are characterized by smooth thermal regions, sparse textures, and weak local contrast, which make faithful structure recovery particularly challenging~\cite{huang2025iisrsurvey,ying2022mocopnet,liang2023dasr}.  
To obtain high-quality results, existing IISR methods explore infrared-specific high-frequency modeling, structural constraints, frequency-domain representations, and powerful generative priors to enhance image details and target boundaries~\cite{huang2021psrgan,ying2022mocopnet,liang2023dasr,li2024corple,fang2024scinet}. 
More recently, pretrained SR and generative models developed on large-scale natural images have also been adapted to infrared imagery, benefiting from their strong restoration and generation capabilities~\cite{wang2024stablesr,wu2024seesr,yu2024supir,wang2024sinsr,wu2024osediff,long2025pftsr,zou2026realiisr}.

Despite their improved visual quality, we observe that sharper textures and stronger high-frequency responses do not necessarily lead to more faithful infrared imaging. Due to the substantial domain discrepancy between visible-light and infrared
imagery~\cite{huang2025iisrsurvey,zou2026realiisr}, cross-domain priors
introduces artificial textures, over-sharpened boundaries,
or spurious high-frequency details. Such visually appealing details alter the original thermal distributions and object contours rather than recover authentic infrared structures~\cite{
li2025difiisr,zou2026realiisr}. More importantly, these structural deviations propagate to downstream perception models by changing object boundaries and semantic cues. As a result, an infrared image that appears sharper to human observers does not necessarily provide a more reliable representation for downstream tasks. This motivates us to reconsider the objective of IISR: rather than merely enhancing perceptual sharpness, IISR should faithfully preserve thermal structures, target contours, and task-relevant semantic information.

To this end, we propose \textbf{FaithIR}, an infrared-domain conditional flow-matching framework operating in pixel space. Different from recent IISR methods~\cite{li2025difiisr,zou2026realiisr},
FaithIR is trained only on paired infrared HR--LR images, without pretrained natural-image generators, separately pretrained image tokenizers, or detector and segmenter supervision, allowing the model to learn infrared-specific thermal distributions and structural characteristics while reducing cross-domain texture bias. Specifically, FaithIR consists of two collaborative branches. The \textbf{patch-level conditioning branch} captures global thermal distributions, object layouts, and long-range structural dependencies from the LR observation, providing structural guidance for restoration. Under this guidance, the \textbf{pixel-level restoration branch} performs dense local reconstruction to recover target contours and fine-grained infrared structures. The entire restoration process is formulated directly in the pixel domain using flow matching~\cite{lipman2023flowmatching,liu2023rectifiedflow,ma2024sit}, enabling structural information to be propagated from global patch representations to local pixel reconstruction without introducing latent image representations~\cite{rombach2022ldm,peebles2023dit,yu2026pixeldit}.

Extensive experiments on benchmarks including FLIR-IISR~\cite{zou2026realiisr}, M3FD~\cite{liu2022m3fd}, and
FMB~\cite{liu2023fmb} achieve strong reconstruction fidelity and robust cross-dataset generalization. More importantly, when fed our SR results into fixed object detection and semantic segmentation, our FaithIR consistently provides more reliable inputs compared with other methods. It demonstrates that faithfully preserving infrared thermal distributions, object contours, and semantic structures is more beneficial to machine perception than simply generating visually sharper high-frequency details.

The main contributions of this work are summarized as follows:
\begin{itemize}
    \item We empirically reveal a discrepancy between perceptual sharpness and task-relevant fidelity in IISR, showing that perceptually favored results do not necessarily better preserve HR infrared structures or downstream predictions.

    \item We propose FaithIR, an infrared-domain pixel-level SR framework, trained without natural-image generator or tokenizer pretraining. IIts patch-level structural conditioning guides dense pixel-level restoration through position-specific pixel-wise AdaLN, reconstructing thermal distributions and target contours.

    \item We conduct extensive experiments covering reconstruction fidelity, cross-dataset generalization, object detection, and semantic segmentation. FaithIR consistently achieves strong reconstruction performance and provides superior inputs for downstream perception, demonstrating the importance of faithful task-relevant reconstruction for downstream tasks.
\end{itemize}

\section{Related Work}

\subsection{Infrared Image Super-Resolution}

Infrared images exhibit smooth thermal distributions, weak local contrast, sparse textures, and sensor-dependent degradations, making IISR fundamentally different from visible-image SR.~\cite{huang2025iisrsurvey}. Early IISR methods improve reconstruction by introducing transfer learning, adversarial learning, motion priors, and attention mechanisms, as exemplified by PSRGAN~\cite{huang2021psrgan}, MoCoPNet~\cite{ying2022mocopnet}, and DASR~\cite{liang2023dasr}. More recent approaches focus on infrared-specific structural and frequency modeling. CoRPLE incorporates contourlet-domain prompts to enhance structural representations~\cite{li2024corple}, InfraFFN combines convolution and attention for effective local--global feature interaction~\cite{qin2025infraffn}, and IRSRMamba introduces wavelet-modulated Mamba blocks for infrared feature reconstruction~\cite{huang2025irsrmamba}. Real-IISR further extends IISR to real-world degradations through thermal-structural autoregression and a condition-adaptive codebook~\cite{zou2026realiisr}. However, most existing IISR methods primarily focus on image-level reconstruction quality, while the utility of recovered details for downstream machine perception remains less explored.

\subsection{Diffusion and Flow-Based Super-Resolution}

Diffusion models have emerged as powerful generative models by progressively transforming noisy samples into clean images through a learned reverse process~\cite{ho2020ddpm,nichol2021iddpm,song2021sde}. Their strong generative capability has motivated extensive applications to image restoration and super-resolution. SR3 formulates super-resolution as a conditional denoising process~\cite{saharia2023sr3}, while subsequent methods improve reconstruction quality and sampling efficiency through different diffusion formulations. ResShift introduces a residual-shifting diffusion process for efficient restoration~\cite{yue2023resshift}, whereas StableSR~\cite{wang2024stablesr} exploits pretrained diffusion priors for high-resolution reconstruction. SinSR further reduces the iterative sampling process to enable efficient single-step super-resolution~\cite{wang2024sinsr}.

More recently, flow-based generative modeling has emerged as an alternative formulation for learning continuous transport between noise and data distributions. Flow matching directly learns a continuous vector field that connects the two distributions~\cite{lipman2023flowmatching}, while rectified flow encourages straighter transport trajectories for more efficient generation~\cite{liu2023rectifiedflow}. SiT further integrates interpolant-based generative objectives with scalable Transformer architectures~\cite{ma2024sit}. These advances highlight the potential of continuous generative formulations and scalable Transformer architectures for efficient and high-fidelity image synthesis.

\begin{figure*}[t]
    \centering
    \includegraphics[width=\textwidth]{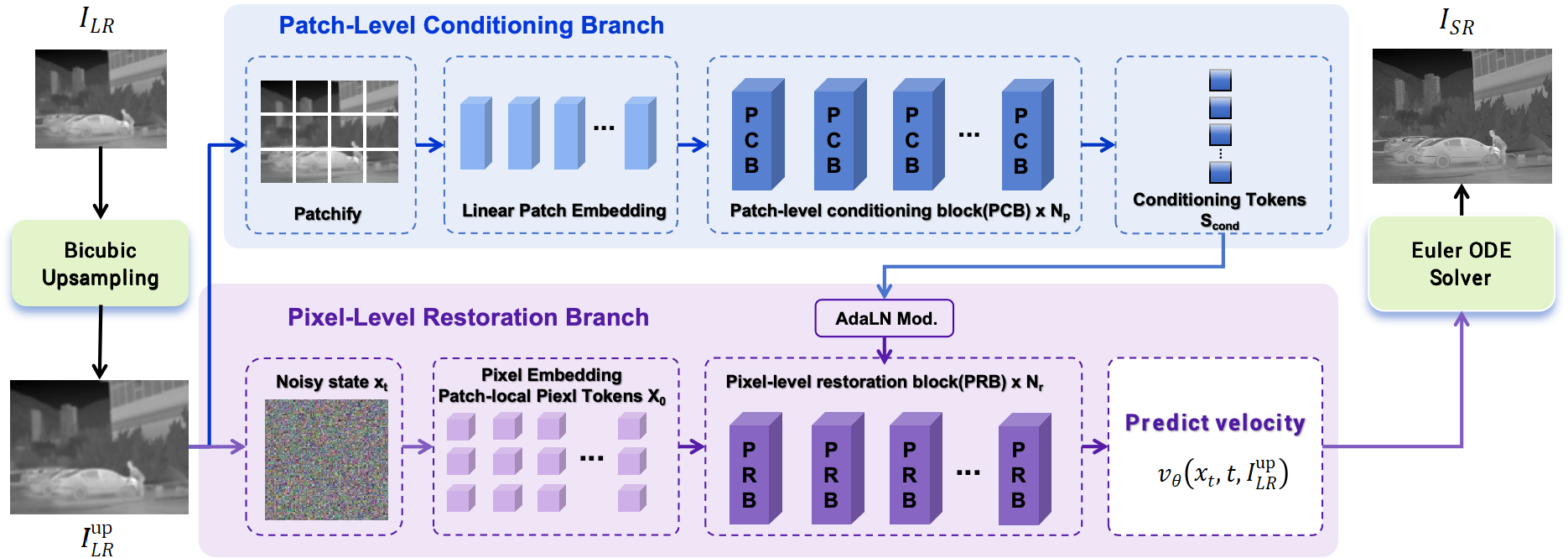}
    \caption{Overview of the proposed FaithIR. 
    % The framework consists of the patch-level conditioning branch for global infrared structural modeling and the pixel-level restoration branch for fine-grained reconstruction in the pixel domain.
    }
    \label{fig:framework}
\end{figure*}

\begin{figure}[t]
    \centering
    \includegraphics[width=\linewidth]{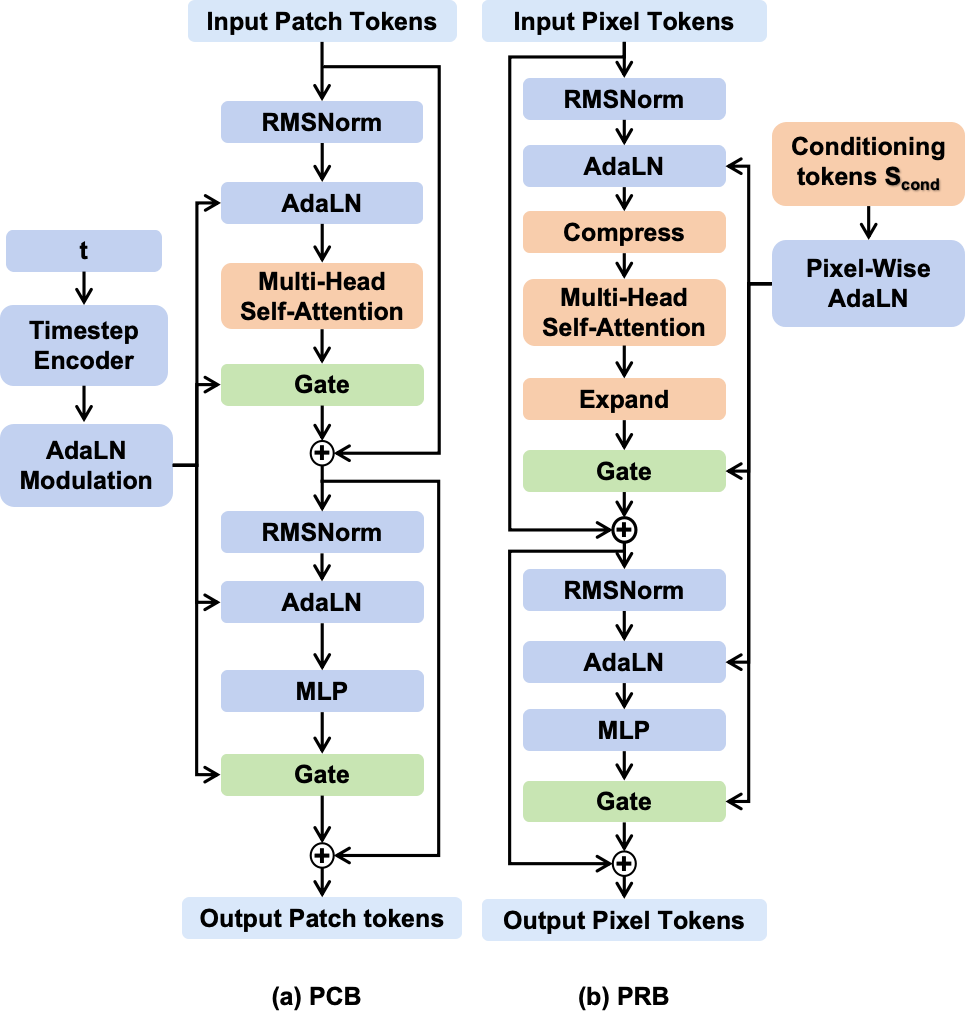}
    \caption{ The proposed (a) Patch-Level Conditioning Block (PCB) and (b) Pixel-Level Restoration Block (PRB).}
    \vspace{-10pt}
    \label{fig:block_architecture}
\end{figure}

\section{Methodology}

\subsection{Overview}

As illustrated in Fig. 1, given a low-resolution infrared image $I_{LR}$, the goal of our FaithIR is to generate its high-resolution counterpart $I_{HR}$  by adopting a dual-branch architecture composed of a Patch-Level Conditioning Branch and a Pixel-Level Restoration Branch. The Patch-Level Conditioning Branch extracts high-level thermal distributions, object layouts, and long-range structural information from the upsampled low-resolution input, providing global guidance for the restoration process. Meanwhile, the Pixel-Level Restoration Branch focuses on dense reconstruction by combining the noisy intermediate representation with the low-resolution observation to recover fine details. By coupling global structural guidance with pixel-level refinement, FaithIR effectively restores infrared-specific thermal patterns and preserves accurate target contours.

\subsection{Patch-Level Conditioning Branch}

Given
$I_{LR}^{\mathrm{up}}\in\mathbb{R}^{B\times C\times H\times W}$,
we divide it into non-overlapping $P\times P$ patches, where
$L=\frac{H}{P}\frac{W}{P}$ denotes the number of patches. The flattened
patches are projected into a $D$-dimensional embedding space:
\begin{equation}
S_0 =
W_{\mathrm{patch}}X_{\mathrm{patch}},
\qquad
S_0\in\mathbb{R}^{B\times L\times D}.
\label{eq:patch_embed}
\end{equation}

The timestep $t$ is encoded into a timestep embedding
$e_t\in\mathbb{R}^{B\times D}$ and injected into each Patch-Level
Conditioning Block (PCB) through adaptive modulation. The stacked PCBs are
written compactly as
\begin{equation}
S_n =
\mathrm{PCB}_n(S_{n-1},e_t),
\qquad n=1,\ldots,N_p.
\label{eq:pcb}
\end{equation}
Each PCB contains RMSNorm, timestep-conditioned AdaLN, multi-head
self-attention, an MLP, and gated residual connections. Two-dimensional
positional information is introduced into self-attention to preserve spatial
relationships among infrared patches.

After $N_p$ PCBs, patch-level features are combined with the timestep
embedding to obtain structural conditioning tokens:
\begin{equation}
S_{\mathrm{cond}}
=
\mathrm{SiLU}(S_{N_p}+e_t),
\qquad
S_{\mathrm{cond}}
\in\mathbb{R}^{B\times L\times D},
\label{eq:scond}
\end{equation}
where $e_t$ is broadcast along the patch dimension. These tokens encode global thermal distributions, object layouts, and long-range structural cues for pixel-level restoration.

\subsection{Pixel-Level Restoration Branch}

The restoration branch directly receives both the noisy state and the
upsampled LR observation. We first concatenate them along the channel
dimension and embed the resulting feature into patch-local pixel tokens:
\begin{equation}
X_0 =
\mathrm{PixelEmbed}
\left([x_t;I_{LR}^{\mathrm{up}}]\right),
\qquad
X_0\in\mathbb{R}^{B\times L\times P^2\times C_p},
\label{eq:pixel_embed}
\end{equation}
where $C_p$ denotes the pixel-token dimension. In contrast to the patch-level branch, each patch retains all $P^2$ pixel tokens, allowing fine-grained local reconstruction.

\paragraph{Pixel-Wise AdaLN Modulation.}
For the $l$-th patch, its conditioning token
$s_l^{\mathrm{cond}}\in\mathbb{R}^{D}$ is projected independently in each
PRB to generate modulation parameters for all $P^2$ pixel positions:
\begin{equation}
M_l^{(m)}
=
W_{\mathrm{AdaLN}}^{(m)}s_l^{\mathrm{cond}}
+b_{\mathrm{AdaLN}}^{(m)}
\in
\mathbb{R}^{P^2\times6C_p}.
\label{eq:pixel_adaln}
\end{equation}
The output is reshaped into six groups of pixel-wise shift, scale, and gate
parameters for the attention and MLP branches. Although pixels within the
same patch share one conditioning token, different pixel positions correspond
to different output channels of the projection layer. Therefore, each pixel
token receives position-specific modulation parameters rather than a single
set of parameters shared across the entire patch.

\paragraph{Compact Global Interaction.}
Performing global attention over all dense pixel tokens is computationally
expensive. We therefore compact the $P^2$ pixel tokens of each patch into one
attention token:
\begin{equation}
\bar{X}_m =
W_{\mathrm{comp}}^{(m)}
\mathrm{vec}(X_{m-1}),
\qquad
\bar{X}_m\in\mathbb{R}^{B\times L\times D_{\mathrm{attn}}}.
\label{eq:compact}
\end{equation}
Global self-attention is performed across the resulting $L$ compact tokens,
after which the features are expanded back to the original $P^2$ pixel-token
layout. In this way, FaithIR enables global information interaction while
maintaining dense pixel-level representations.

The restoration branch stacks $N_r$ PRBs, each combining pixel-wise
conditioning, compact global interaction, and local MLP refinement:
\begin{equation}
X_m=\mathrm{PRB}_m(X_{m-1},S_{\mathrm{cond}}),
\quad m=1,\ldots,N_r.
\label{eq:prb}
\end{equation}
After the final PRB, the output tokens $X_{N_r}$ are projected by the
prediction head to obtain the velocity field
$v_\theta=\mathrm{Head}(X_{N_r})$, which has the same spatial resolution
as the target HR image.

\subsection{Conditional Flow Matching Objective}

Given the ground-truth HR image $I_{HR}$ and Gaussian noise
$\epsilon\sim\mathcal{N}(0,I)$, we sample $t\in(0,1)$ and construct the
intermediate state as
\begin{equation}
x_t =
\alpha_t I_{HR}+\sigma_t\epsilon,
\qquad
\alpha_t=t,\quad\sigma_t=1-t.
\label{eq:flow_path}
\end{equation}

The corresponding target velocity field is
\begin{equation}
v^{*}
=
\frac{d x_t}{dt}
=
I_{HR}-\epsilon.
\label{eq:target_velocity}
\end{equation}

FaithIR is optimized by matching the predicted velocity to the target field:
\begin{equation}
\mathcal{L}_{\mathrm{FM}}
=
\mathbb{E}
\left[
\left\|
v_\theta(x_t,t,I_{LR}^{\mathrm{up}})
-
(I_{HR}-\epsilon)
\right\|_2^2
\right].
\label{eq:fm_loss}
\end{equation}

During inference, we initialize $x_0\sim\mathcal{N}(0,I)$ and integrate the
learned vector field from $t=0$ to $t=1$. Using Euler integration, the update
at each step is
\begin{equation}
x_{t_{k+1}}
=
x_{t_k}
+
(t_{k+1}-t_k)
v_\theta(x_{t_k},t_k,I_{LR}^{\mathrm{up}}),
\label{eq:euler}
\end{equation}
and the final state $x_1$ is taken as the super-resolved infrared image
$I_{SR}$.

\section{Experiments}

\begin{table*}[t]
\centering
\footnotesize
\renewcommand{\arraystretch}{1.05}

\begin{tabular*}{\textwidth}
{@{\extracolsep{\fill}}c c | c c c c | c c | c@{}}
\hline

\multirow{2}{*}{\textbf{Method}} &
\multirow{2}{*}{\textbf{Type}} &
\multicolumn{4}{c|}{\textbf{SR Metric}} &
\multicolumn{2}{c|}{\textbf{Object Detection}} &
\multicolumn{1}{c}{\textbf{Segmentation}} \\

\cline{3-9}

& &
\textbf{PSNR} $\uparrow$ &
\textbf{SSIM} $\uparrow$ &
\textbf{LPIPS} $\downarrow$ &
\textbf{FID} $\downarrow$ &
\textbf{Det. Num.} $\uparrow$ &
\textbf{Det. Rate (\%)} $\uparrow$ &
\textbf{Seg. Dev. (\%)} $\downarrow$ \\
\hline

PFT-SR (CVPR'25) &
ISR &
23.67 &
\underline{0.8003} &
0.4125 &
101.02 &
420 &
61.90 &
10.86 \\

SinSR (CVPR'24) &
ISR &
22.93 &
0.7553 &
0.4049 &
108.81 &
461 &
69.10 &
11.92 \\

HAT (CVPR'23) &
ISR &
23.72 &
0.7973 &
0.4011 &
100.08 &
476 &
70.60 &
10.54 \\

InfraFFN (KBS'25) &
IISR &
\underline{23.84} &
0.7982 &
0.4093 &
104.84 &
\underline{491} &
\underline{71.70} &
11.48 \\

DifIISR (CVPR'25) &
IISR &
22.95 &
0.7196 &
0.4860 &
110.96 &
446 &
66.80 &
12.68 \\

CoRPLE (ECCV'24) &
IISR &
23.80 &
0.7949 &
0.4113 &
110.11 &
481 &
69.40 &
11.17 \\

Real-IISR (CVPR'26) &
IISR &
21.36 &
0.7471 &
\textbf{0.2765} &
\underline{81.58} &
412 &
63.40 &
\underline{4.88} \\

\hline

\textbf{FaithIR (Ours)} &
IISR &
\textbf{24.91} &
\textbf{0.8121} &
\underline{0.2813} &
\textbf{74.28} &
\textbf{594} &
\textbf{72.10} &
\textbf{3.66} \\

\hline
\end{tabular*}

\caption{Quantitative comparison on the FLIR-IISR test set.}
\label{tab:flir_comparison}
\end{table*}

% =========================================================
% 4.1 Experimental Settings
% =========================================================
\subsection{Experimental Settings}

\paragraph{Implementation Details.}

FaithIR is trained from scratch on FLIR-IISR using four NVIDIA A100 GPUs, without relying on pretrained natural-image generative models or image tokenizers. Spatially aligned  64×64 LR  and 256×256 HR patches are randomly cropped for training, where the LR input is bicubically upsampled to the HR resolution. FaithIR is optimized using AdamW with the conditional flow-matching objective. More  details are provided in Appendix~\ref{app:training_details}.
% \NAX{ Please refer toAppendix~\ref{supp-app:training_details} for more training details.}Additional qualitative comparisons are provided in Appendix~\ref{app:training_details}.

\paragraph{Datasets.}
We train and evaluate all SR methods on FLIR-IISR~\cite{zou2026realiisr}, which contains 1,457 paired infrared images. We use 1,192 pairs for training and the remaining 265 pairs for testing. The native HR and LR resolutions are $1024\times768$ and $256\times192$, respectively. We further evaluate cross-dataset generalization on M3FD~\cite{liu2022m3fd} and FMB~\cite{liu2023fmb}. For both datasets, the original infrared images are treated as HR references, while the corresponding LR inputs are generated by
$4\times$ bicubic downsampling. All SR models are trained only on FLIR-IISR and directly evaluated on M3FD and FMB without any fine-tuning. M3FD and FMB are additionally used for ground-truth object detection and semantic segmentation evaluation, respectively.

\paragraph{Metrics.} 
% For SR evaluation, we use PSNR and SSIM~\cite{wang2004ssim} to measure reconstruction fidelity and LPIPS~\cite{zhang2018lpips} to evaluate full-reference perceptual similarity. FID~\cite{heusel2017fid} is additionally reported to characterize the feature-distribution discrepancy between    reconstructed and HR images. We also report CLIPIQA~\cite{wang2023clipiqa}, MANIQA~\cite{yang2022maniqa}, MUSIQ~\cite{ke2021musiq}, and NIQE~\cite{mittal2013niqe} as complementary no-reference quality metrics. Higher PSNR, SSIM, CLIPIQA, MANIQA, and MUSIQ values are preferred, whereas lower LPIPS, FID, and NIQE values indicate
% better performance. \NAX {Object detection, Segmentation}
For SR evaluation, we use PSNR and SSIM~\cite{wang2004ssim} to measure reconstruction fidelity, and LPIPS~\cite{zhang2018lpips} to evaluate full-reference perceptual similarity. FID~\cite{heusel2017fid} is
additionally reported to characterize the feature-distribution discrepancy between reconstructed and HR images. We also report CLIPIQA~\cite{wang2023clipiqa}, MANIQA~\cite{yang2022maniqa}, MUSIQ~\cite{ke2021musiq}, and NIQE~\cite{mittal2013niqe} as complementary no-reference quality metrics.
For downstream evaluation, we report Det. Num., Det. Rate, and Seg. Dev. on FLIR-IISR. Object detection on M3FD is evaluated using $\mathrm{mAP}_{50}$, and class-wise recall, while semantic segmentation on FMB is evaluated using mIoU and class-wise IoU.

\paragraph{Comparative Methods.}

We compare FaithIR with representative general ISR
methods, including PFT-SR~\cite{long2025pftsr}, HAT~\cite{chen2023hat},
and SinSR~\cite{wang2024sinsr}, and IISR
methods, including CoRPLE~\cite{li2024corple},
InfraFFN~\cite{qin2025infraffn}, DifIISR~\cite{li2025difiisr}, and
Real-IISR~\cite{zou2026realiisr}. Among them, Real-IISR, DifIISR, SinSR,
and PFT-SR leverage pretrained priors learned from natural visible-light
images, whereas HAT, CoRPLE, and InfraFFN are trained from scratch on infrared
data. These methods cover both transfer-based and infrared-domain learning
paradigms, enabling comprehensive comparison in terms of reconstruction
quality, perceptual characteristics, cross-dataset generalization, and
task-related information preservation.

% =========================================================
% 4.2 Image Reconstruction Evaluation
% =========================================================
% \NAX{Experimental results}

% \NAX{Quantitative results for image super-resolution}

% \NAX{Qualitative results for image super-resolution}

% \NAX{Quantitative results for Object Detectio}

% \NAX{Qualitative results for Object Detectio}

\subsection{Image Super-Resolution Evaluation}

\begin{table}[t]
\centering
\footnotesize
\setlength{\tabcolsep}{2.5pt}
\renewcommand{\arraystretch}{1.05}

\begin{tabular}{c|cccc}
\hline
\textbf{Method} &
\textbf{CLIPIQA} $\uparrow$ &
\textbf{MANIQA} $\uparrow$ &
\textbf{MUSIQ} $\uparrow$ &
\textbf{NIQE} $\downarrow$ \\
\hline

\multicolumn{5}{c}{\textit{With Natural-Image Pretrained Prior}} \\
\hline
PFT-SR
& 0.2458
& 0.2396
& 35.14
& 9.28 \\

SinSR
& \underline{0.4896}
& \textbf{0.3224}
& \textbf{50.64}
& 7.41 \\

DifIISR
& \textbf{0.5046}
& \underline{0.3207}
& 49.53
& 8.25 \\

Real-IISR
& 0.1987
& 0.2782
& \underline{49.81}
& \underline{6.83} \\

\hline
\multicolumn{5}{c}{\textit{Without Natural-Image Pretrained Prior}} \\
\hline
HAT
& 0.2834
& 0.2352
& 29.50
& 9.74 \\

InfraFFN
& 0.2646
& 0.2338
& 29.63
& 9.84 \\

CoRPLE
& 0.1795
& 0.2462
& 29.39
& 8.95 \\

\textbf{FaithIR (Ours)}
& 0.2387
& 0.2420
& 37.11
& \textbf{6.32} \\

\hline
\end{tabular}
\caption{No-reference quality comparison on FLIR-IISR.}
% \vspace{-12pt}
\label{tab:nr_comparison}
\end{table}

\begin{table}[t]
\centering
\footnotesize
\renewcommand{\arraystretch}{1.05}

\begin{tabular*}{0.95\columnwidth}
{@{\extracolsep{\fill}}c|cccc@{}}
\hline
\textbf{Method} &
\textbf{PSNR}$\uparrow$ &
\textbf{SSIM}$\uparrow$ &
\textbf{LPIPS}$\downarrow$ &
\textbf{FID}$\downarrow$ \\
\hline

PFT-SR      & 29.88 & 0.9119 & 0.1968 & 32.16 \\
SinSR       & 31.71 & 0.8718 & 0.2809 & 51.27 \\
HAT         & \underline{33.34} & \textbf{0.9382} & \underline{0.1750} & 24.49 \\
InfraFFN    & 32.84 & 0.9253 & 0.2138 & 42.47 \\
DifIISR     & 32.02 & 0.9225 & 0.2608 & 90.32 \\
CoRPLE      & 33.32 & \underline{0.9347} & 0.1828 & \underline{22.69} \\
Real-IISR   & 25.95 & 0.8462 & 0.2400 & 30.84 \\
\hline

\textbf{FaithIR (Ours)}
& \textbf{33.92}
& 0.9259
& \textbf{0.1501}
& \textbf{21.11} \\

\hline
\end{tabular*}

\caption{Quantitative results on M3FD.}
% \vspace{-10pt}
\label{tab:m3fd_comparison}
\end{table}
\begin{table}[t]
\centering
\footnotesize
\renewcommand{\arraystretch}{1.05}

\begin{tabular*}{0.95\columnwidth}
{@{\extracolsep{\fill}}c|cccc@{}}
\hline
\textbf{Method} &
\textbf{PSNR}$\uparrow$ &
\textbf{SSIM}$\uparrow$ &
\textbf{LPIPS}$\downarrow$ &
\textbf{FID}$\downarrow$ \\
\hline

PFT-SR      & 37.59 & \textbf{0.9680} & \underline{0.1756} & \underline{37.82} \\
SinSR       & 35.16 & 0.9269 & 0.2401 & 82.01 \\
HAT         & 37.55 & \underline{0.9636} & 0.2026 & 39.45 \\
InfraFFN    & \textbf{37.98} & 0.9608 & 0.2091 & 43.64 \\
DifIISR     & 37.48 & 0.9619 & 0.2210 & 109.09 \\
CoRPLE      & 37.15 & 0.9608 & 0.2283 & 43.04 \\
Real-IISR   & 26.53 & 0.8821 & 0.2589 & 79.37 \\
\hline

\textbf{FaithIR (Ours)}
& \underline{37.64}
& 0.9576
& \textbf{0.1595}
& \textbf{33.62} \\

\hline
\end{tabular*}

\caption{Quantitative results on FMB.}
% \vspace{-26pt}
\label{tab:fmb_comparison}
\end{table}
% Tables~\ref{tab:m3fd_comparison} and~\ref{tab:fmb_comparison} report
% cross-dataset reconstruction results on M3FD and FMB, respectively.
% On M3FD, FaithIR achieves the best PSNR, LPIPS, and FID while maintaining
% competitive SSIM. On FMB, FaithIR achieves the best LPIPS
% and FID while remaining competitive in PSNR and SSIM. The consistent
% performance across the two unseen datasets demonstrates that FaithIR maintains
% robust reconstruction capability under changes in scene content and data
% distribution, supporting the effectiveness of direct infrared-domain learning
% for cross-dataset super-resolution.

% \paragraph{Qualitative Reconstruction Comparison.}

\begin{figure*}[!t]
    \centering
    \includegraphics[width=\textwidth]{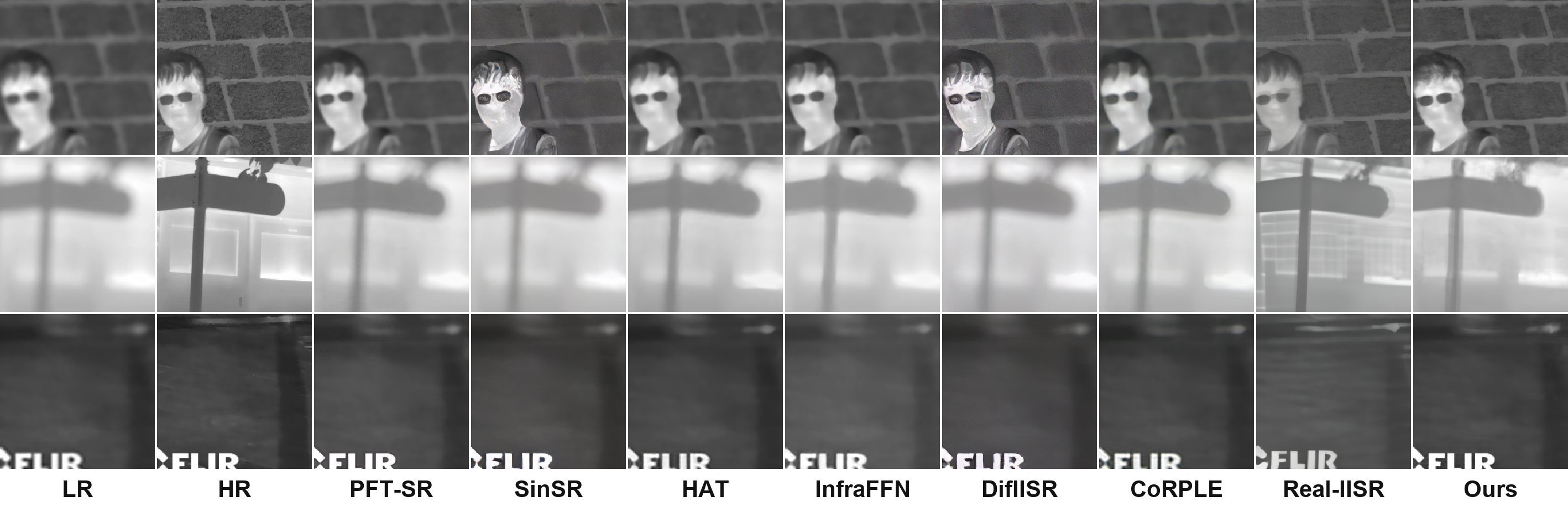}
    \caption{
    Qualitative comparison on the FLIR-IISR test set.
    }
    \label{fig:qualitative}
\end{figure*}
\begin{figure*}[t]
    \centering
    \includegraphics[width=\textwidth]{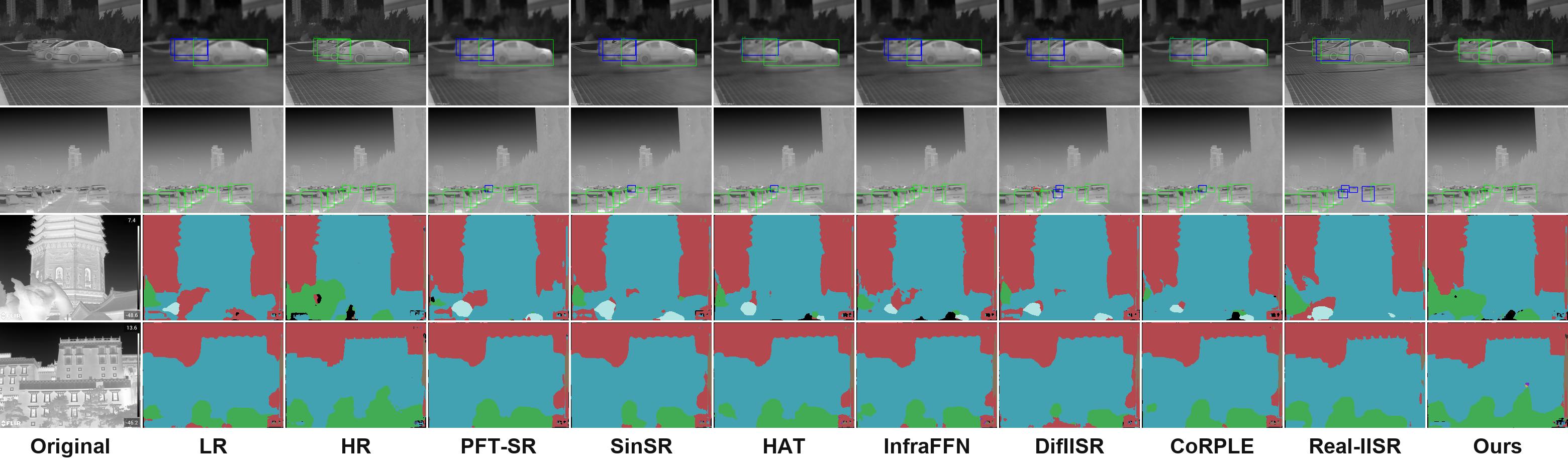}
    \caption{Downstream task results on the FLIR-IISR test set. The first two rows show object detection, and the last two rows show semantic segmentation predictions using the same fixed downstream models.}
    \label{fig:flir_downstream}
\end{figure*}

\paragraph{Quantitative Results.}

As shown in Table~\ref{tab:flir_comparison}, FaithIR achieves competitive SR metrics, including PSNR, SSIM, FID, and LPIPS, demonstrating strong reconstruction fidelity and distributional consistency with real HR infrared observations. Table~\ref{tab:nr_comparison} further reports no-reference image quality metrics~\cite{mittal2013niqe,ke2021musiq,yang2022maniqa,wang2023clipiqa}. Although pretrained-prior-based methods, including PFT-SR, SinSR, DifIISR, and Real-IISR, achieve higher scores on some no-reference metrics such as CLIPIQA, MANIQA, or MUSIQ, their performance is less competitive in full-reference super-resolution quality and downstream perception tasks. Tables~\ref{tab:m3fd_comparison} and \ref{tab:fmb_comparison} report cross-dataset reconstruction results on M3FD and FMB, respectively. Without any fine-tuning on either dataset, FaithIR maintains strong reconstruction performance across multiple image-level reconstruction metrics. These results demonstrate that the learned infrared-domain representation generalizes robustly to unseen scenes and data distributions.

\paragraph{Qualitative Results.}
% Figure~\ref{fig:qualitative} further compares the visual results
% of different methods on FLIR-IISR\NAX{Please refer to Appendix~\ref{} for more visual results on other datasets.}.
% FaithIR preserves clear target contours and coherent local thermal structures
% while maintaining smooth infrared intensity distributions.
% In contrast, some competing methods either over-smooth structural details or
% introduce visually sharp high-frequency patterns that deviate from the
% corresponding HR observations.
% These comparisons demonstrate that FaithIR better balances local detail recovery
% and structural fidelity.
Figure~\ref{fig:qualitative} shows a qualitative comparison of different methods on the FLIR-IISR test set. FaithIR preserves a clear target profile and coherent local thermal structure while maintaining a smooth infrared intensity distribution. In contrast, some competing methods either over-smooth structural details or introduce visually overly sharp high-frequency details, deviating from the corresponding high-resolution observations. These results demonstrate that FaithIR achieves a better balance between local detail recovery and faithful preservation of infrared structure. Please refer to Appendix~\ref{app:img_SR} for more visual results on other datasets.

\begin{figure*}[t]
    \centering
    \includegraphics[width=\textwidth]{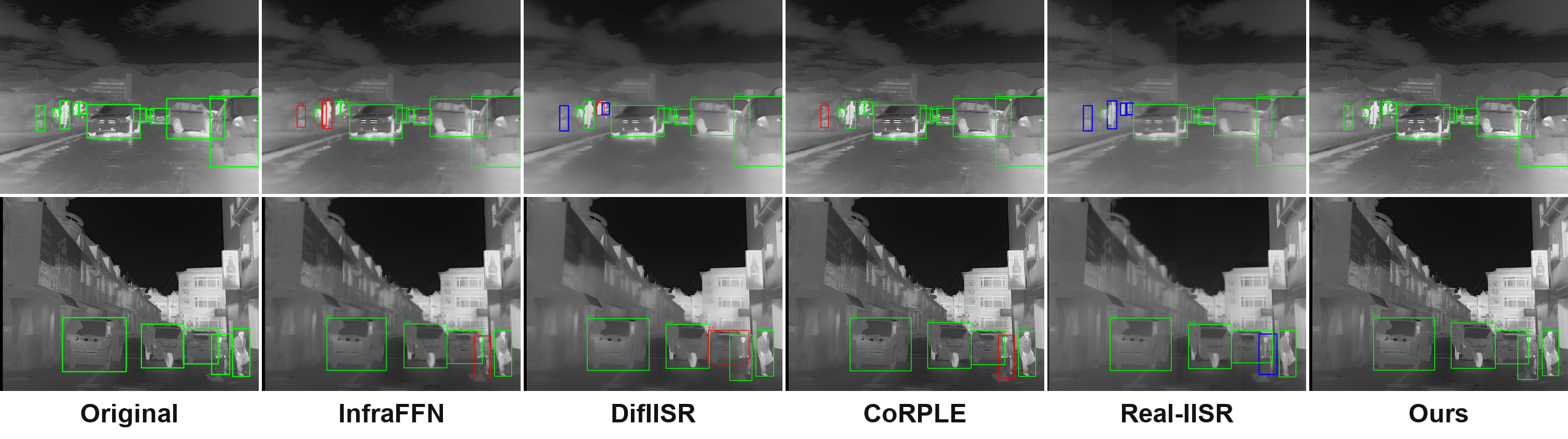}
    \caption{Object detection results on M3FD using the same fixed detector. Green, red, and blue boxes indicate correct detections, false positives, and false negatives, respectively.
    }
    % \vspace{-8pt}
    \label{fig:m3fd_detection}
\end{figure*}
\begin{figure*}[t]
    \centering
    \includegraphics[width=\textwidth]{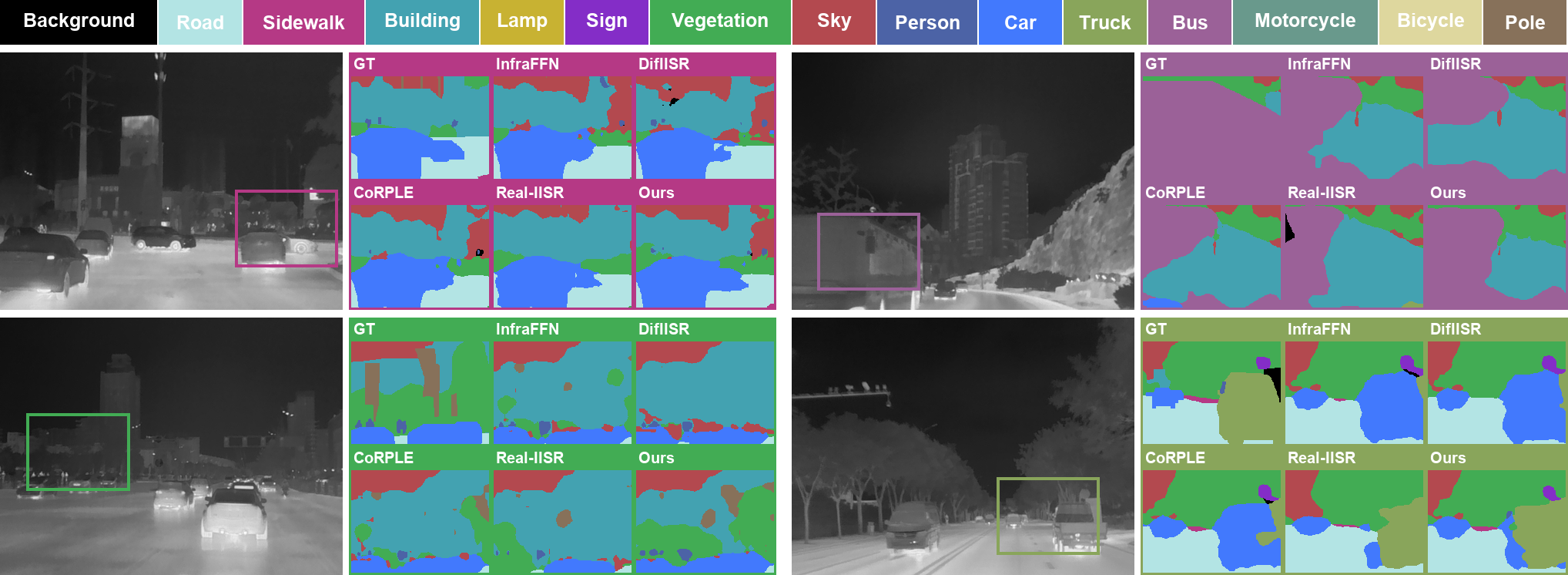}
    \caption{
    Semantic segmentation results on FMB using the same fixed segmentation model. Highlighted regions are enlarged for comparison with the ground-truth labels.
    }
    \label{fig:fmb_segmentation}
\end{figure*}
% \paragraph{Detection on M3FD and FLIR-IISR.\NAX{!!!!!!!!!!}}
% To further validate the downstream benefit using ground-truth annotations,
% we evaluate object detection on M3FD.
% We employ YOLOv8n as the detector and keep it fixed for all SR
% methods~\cite{jocher2023yolov8}.
% The original M3FD infrared images are treated as HR references, and the
% corresponding LR inputs are generated by $4\times$ bicubic downsampling.
% We report Det. Num., $\mathrm{mAP}_{50}$, and class-wise recall.
% Detailed detector training configurations are provided in the supplementary
% material.
% \begin{table}[t]
% \centering
% \small
% \setlength{\tabcolsep}{3pt}
% \renewcommand{\arraystretch}{1.05}

% \begin{tabular}{c | c c c c c}
% \hline

% \multirow{2}{*}{\textbf{Method}} &
% \multicolumn{5}{c}{\textbf{Object Detection}} \\

% \cline{2-6}

% &
% \textbf{mAP$_{50}$} $\uparrow$ &
% \textbf{Person} $\uparrow$ &
% \textbf{Car} $\uparrow$ &
% \textbf{Lamp} $\uparrow$ &
% \textbf{Motor.} $\uparrow$ \\
% \hline

% InfraFFN &
% 0.643 & 0.727 & 0.694 & 0.253 & 0.414 \\

% DifIISR &
% 0.602 & 0.660 & 0.686 & 0.124 & 0.357 \\

% CoRPLE &
% 0.656 & 0.730 & 0.719 & 0.291 & 0.471 \\

% Real-IISR &
% 0.495 & 0.462 & 0.582 & 0.104 & 0.271 \\

% \hline

% \textbf{FaithIR (Ours)} &
% \textbf{0.661} &
% \textbf{0.732} &
% \textbf{0.724} &
% \textbf{0.324} &
% \textbf{0.514} \\

% \hline
% \end{tabular}

% \caption{Object detection results on M3FD.}
% \label{tab:detection_m3fd}
% \end{table}
\begin{table}[t]
\centering
\small
\setlength{\tabcolsep}{3pt}
\renewcommand{\arraystretch}{1.05}

\begin{tabular}{c | c c c c c}
\hline

\multirow{2}{*}{\textbf{Method}} &
\multicolumn{5}{c}{\textbf{Object Detection}} \\

\cline{2-6}

&
\textbf{mAP$_{50}$} $\uparrow$ &
\textbf{Person} $\uparrow$ &
\textbf{Car} $\uparrow$ &
\textbf{Lamp} $\uparrow$ &
\textbf{Motor.} $\uparrow$ \\
\hline

InfraFFN &
0.643 & 0.727 & 0.694 & 0.253 & 0.414 \\

DifIISR &
0.602 & 0.660 & 0.686 & 0.124 & 0.357 \\

CoRPLE &
\underline{0.656} &
\underline{0.730} &
\underline{0.719} &
\underline{0.291} &
\underline{0.471} \\

Real-IISR &
0.495 & 0.462 & 0.582 & 0.104 & 0.271 \\

\hline

\textbf{FaithIR (Ours)} &
\textbf{0.661} &
\textbf{0.732} &
\textbf{0.724} &
\textbf{0.324} &
\textbf{0.514} \\

\hline
\end{tabular}

\caption{Object detection results on M3FD.}
% \vspace{-10pt}
\label{tab:detection_m3fd}
\end{table}
\begin{table}[t]
\centering
\small
\setlength{\tabcolsep}{3pt}
\renewcommand{\arraystretch}{1.05}

\begin{tabular}{c | c c c c c}
\hline

\multirow{2}{*}{\textbf{Method}} &
\multicolumn{5}{c}{\textbf{Semantic Segmentation}} \\

\cline{2-6}

&
\textbf{mIoU} $\uparrow$ &
\textbf{Pole} $\uparrow$ &
\textbf{Sidew.} $\uparrow$ &
\textbf{Build.} $\uparrow$ &
\textbf{Motor.} $\uparrow$ \\
\hline

DifIISR &
44.4 & 15.9 & 40.0 & 69.6 & 30.1 \\

Real-IISR &
43.5 & 15.0 & 36.7 & 72.3 & 22.6 \\

InfraFFN &
47.6 &
\underline{17.3} &
40.7 &
74.4 &
30.4 \\

CoRPLE &
\underline{48.6} &
16.9 &
\underline{42.7} &
\underline{75.6} &
\underline{32.5} \\

\hline

\textbf{FaithIR (Ours)} &
\textbf{49.0} &
\textbf{19.1} &
\textbf{43.2} &
\textbf{76.4} &
\textbf{33.1} \\

\hline
\end{tabular}

\caption{Semantic segmentation results on FMB.}
\label{tab:segmentation_fmb}
\end{table}
% \begin{table*}[t]
% \centering

% \setlength{\tabcolsep}{5pt}
% \renewcommand{\arraystretch}{1.05}

% \resizebox{0.9\textwidth}{!}{
% \begin{tabular}{c | c c c c c c c c}
% \hline
% \textbf{Method} &
% \textbf{mIoU}$\uparrow$ &
% \textbf{PA}$\uparrow$ &
% \textbf{Pole}$\uparrow$ &
% \textbf{Sidewalk}$\uparrow$ &
% \textbf{Building}$\uparrow$ &
% \textbf{Person}$\uparrow$ &
% \textbf{Car}$\uparrow$ &
% \textbf{Motorcycle}$\uparrow$ \\
% \hline

% \multicolumn{9}{c}{\textit{With Natural-Image Pretrained Prior}} \\
% \hline

% DifIISR
% & 44.4
% & 83.1
% & 15.9
% & 40.0
% & 69.6
% & 55.0
% & 72.2
% & 30.1 \\

% Real-IISR
% & 43.5
% & 84.7
% & 15.0
% & 36.7
% & 72.3
% & 33.0
% & 71.0
% & 22.6 \\

% \hline
% \multicolumn{9}{c}{\textit{Without Natural-Image Pretrained Prior}} \\
% \hline

% InfraFFN
% & 47.6
% & 86.6
% & 17.3
% & 40.7
% & 74.4
% & 58.3
% & 74.8
% & 30.4 \\

% CoRPLE
% & 48.6
% & 86.9
% & 16.9
% & 42.7
% & 75.6
% & \textbf{59.7}
% & \textbf{75.1}
% & 32.5 \\

% \textbf{FaithIR (Ours)}
% & \textbf{49.0}
% & \textbf{87.6}
% & \textbf{19.1}
% & \textbf{43.2}
% & \textbf{76.4}
% & 57.6
% & 74.5
% & \textbf{33.1} \\

% \hline
% \end{tabular}
% }

% \caption{Semantic segmentation results on FMB.}
% \label{tab:segmentation_fmb}

% \end{table*}

\subsection{Downstream Task Evaluation}

% \begin{table*}[t]
% \centering

% \setlength{\tabcolsep}{7pt}
% \renewcommand{\arraystretch}{1.05}

% \begin{tabular}{c | c c c c c c c c}
% \hline
% \textbf{Method} &
% \textbf{mIoU}$\uparrow$ &
% \textbf{PA}$\uparrow$ &
% \textbf{Pole}$\uparrow$ &
% \textbf{Sidewalk}$\uparrow$ &
% \textbf{Building}$\uparrow$ &
% \textbf{Person}$\uparrow$ &
% \textbf{Car}$\uparrow$ &
% \textbf{Motorcycle}$\uparrow$ \\
% \hline

% \multicolumn{9}{c}{\textit{With Natural-Image Pretrained Prior}} \\
% \hline

% DifIISR
% & 44.4
% & 83.1
% & 15.9
% & 40.0
% & 69.6
% & 55.0
% & 72.2
% & 30.1 \\

% Real-IISR
% & 43.5
% & 84.7
% & 15.0
% & 36.7
% & 72.3
% & 33.0
% & 71.0
% & 22.6 \\

% \hline
% \multicolumn{9}{c}{\textit{Without Natural-Image Pretrained Prior}} \\
% \hline

% InfraFFN
% & 47.6
% & 86.6
% & 17.3
% & 40.7
% & 74.4
% & 58.3
% & 74.8
% & 30.4 \\

% CoRPLE
% & 48.6
% & 86.9
% & 16.9
% & 42.7
% & 75.6
% & \textbf{59.7}
% & \textbf{75.1}
% & 32.5 \\

% \textbf{FaithIR (Ours)}
% & \textbf{49.0}
% & \textbf{87.6}
% & \textbf{19.1}
% & \textbf{43.2}
% & \textbf{76.4}
% & 57.6
% & 74.5
% & \textbf{33.1} \\

% \hline
% \end{tabular}

% \caption{Semantic segmentation results on FMB.}
% \label{tab:segmentation_fmb}

% \end{table*}
% \subsection{Downstream Task Evaluation}

The preceding SR results show that FaithIR better preserves infrared
structures at the image level. We further investigate whether such fidelity improvements translate into more reliable inputs for downstream  perception tasks  through object detection and semantic segmentation.

\paragraph{Object Detection on FLIR-IISR and M3FD.}

For FLIR-IISR, which lacks ground-truth detection annotations, we apply a fixed detector to SR and HR images and evaluate their prediction consistency using Det. Num. and Det. Rate (Appendix~\ref{FLIR-IISR-metrics}).
As shown in Table~\ref{tab:flir_comparison}, FaithIR achieves the most favorable detection statistics.

We further evaluate ground-truth detection on M3FD using a fixed YOLOv8n~\cite{jocher2023yolov8}.
Table~\ref{tab:detection_m3fd} shows that FaithIR achieves the best overall $\mathrm{mAP}_{50}$ and strong class-wise recall, while Figure~\ref{fig:m3fd_detection} shows fewer false positives and missed detections. These results demonstrate better preservation of task-relevant infrared structures. Additional quantitative and qualitative detection results are provided in Appendix~\ref{app:detection}.

\paragraph{Semantic Segmentation on FLIR-IISR and FMB.}

For FLIR-IISR, which lacks ground-truth segmentation annotations, we use Seg. Dev. to measure the semantic deviation between predictions on SR images and the corresponding HR references, with its definition provided in Appendix~\ref{FLIR-IISR-metrics}. As shown in Table~\ref{tab:flir_comparison}, FaithIR achieves the lowest semantic deviation, indicating better preservation of task-relevant semantic information.

We further conduct ground-truth semantic segmentation evaluation on FMB using a fixed SegFormer-B2~\cite{xie2021segformer}. Table~\ref{tab:segmentation_fmb} shows that FaithIR achieves the highest mIoU and strong class-wise IoU performance, while Figure~\ref{fig:fmb_segmentation} demonstrates more accurate semantic boundaries and small regions.
Additional quantitative and qualitative results are provided in Appendix~\ref{app:segmentation}.
\subsection{Ablation Study}
 We conduct experiments to evaluate pixel-domain restoration, the Patch-Level Conditioning Branch, and pixel-wise AdaLN across reconstruction and downstream tasks.

We first investigate the choice of restoration space. As shown in Table~\ref{tab:ablation}(a), the pixel-domain formulation consistently outperforms the latent-space DiT baseline in terms of PSNR, SSIM, LPIPS, and FID. Table~\ref{tab:ablation}(b) further shows that pixel-domain restoration also leads to better object detection and semantic segmentation performance. These results suggest that direct restoration in the pixel domain is better suited to preserving fine-grained thermal distributions and target structures, while avoiding the information loss introduced by an additional latent encoding--decoding process. The consistent improvements further demonstrate the benefit of preserving pixel-level infrared information for downstream perception tasks.
\begin{table}[!t]
\centering

\small
\renewcommand{\arraystretch}{1.05}

\begin{minipage}{0.96\columnwidth}
\centering

% ===================== (a) =====================
\textbf{(a) SR results on FLIR-IISR}

\vspace{1pt}

\begin{tabular*}{\linewidth}
{@{\extracolsep{\fill}}lcccc@{}}
\hline
\textbf{Method}
& \textbf{PSNR}$\uparrow$
& \textbf{SSIM}$\uparrow$
& \textbf{LPIPS}$\downarrow$
& \textbf{FID}$\downarrow$ \\
\hline

Latent-Space DiT
& 24.70 & 0.8096 & 0.3023 & 95.26 \\

w/o Cond. Branch
& 23.74 & 0.8031 & 0.3220 & 102.86 \\

Patch-Shared AdaLN
& 24.67 & 0.8073 & 0.2842 & 86.04 \\

\textbf{FaithIR (Full)}
& \textbf{24.91}
& \textbf{0.8121}
& \textbf{0.2813}
& \textbf{74.28} \\

\hline
\end{tabular*}

\vspace{5pt}

% ===================== (b) =====================
\textbf{(b) Detection on M3FD and Segmentation on FMB}

\vspace{1pt}

\begin{tabular*}{\linewidth}
{@{\extracolsep{\fill}}lcc@{}}
\hline
\textbf{Method}
& \textbf{Detection}
& \textbf{Segmentation} \\
& $\mathrm{mAP}_{50}\uparrow$
& $\mathrm{mIoU}\uparrow$ \\
\hline

Latent-Space DiT
& 0.630 & 47.9 \\

w/o Cond. Branch
& 0.630 & 47.6 \\

Patch-Shared AdaLN
& 0.642 & 48.3 \\

\textbf{FaithIR (Full)}
& \textbf{0.661}
& \textbf{49.0} \\

\hline
\end{tabular*}

\caption{Ablation study of FaithIR.}
\label{tab:ablation}

\end{minipage}
\end{table}

% We first investigate the choice of restoration space. As shown in Table~\ref{tab:ablation}(a), the pixel-domain formulation consistently outperforms the latent-space DiT baseline in terms of PSNR, SSIM, LPIPS,and FID. Table~\ref{tab:ablation}(b) further shows that pixel-domain restoration also leads to better object detection and semantic segmentation performance. These results suggest that direct restoration in the pixel domain is better suited to preserving fine-grained thermal distributions and target structures, while avoiding the information loss introduced by an additional latent encoding--decoding process. The consistent improvements in downstream tasks further demonstrate the benefit of preserving pixel-level infrared information for machine perception\NAX{!!!!!!!!!}.

We next examine the role of the Patch-Level Conditioning Branch. Removing this branch causes clear degradation in both reconstruction quality and downstream performance, indicating that global thermal distributions, object layouts, and long-range structural information provide important guidance for fine-grained pixel-level restoration.

Finally, we replace pixel-wise AdaLN with patch-shared modulation, where all pixel positions within the same patch share the same modulation parameters. This variant consistently degrades both reconstruction and downstream performance, demonstrating the importance of position-specific conditioning. By assigning different modulation parameters to individual pixel positions, pixel-wise AdaLN enables finer adaptation to local thermal and structural variations within each patch. Please refer to Appendix~\ref{app:ablation study} for visual results.

\section{Conclusion}

This study investigates infrared image super-resolution from the perspectives of faithful reconstruction and reliable downstream tasks. Our results indicate that visually sharper reconstructions do not necessarily preserve the thermal distributions and structural characteristics of infrared images and may provide limited benefits for downstream perception tasks. To address this issue, we propose FaithIR, a pixel-level infrared super-resolution framework integrating patch-level structural conditioning with dense pixel-level reconstruction. By learning from infrared HR--LR pairs, FaithIR better preserves infrared-specific thermal structures, target contours, and semantic information while reducing reliance on natural-image priors. Extensive experiments demonstrate strong reconstruction fidelity and cross-dataset generalization, while consistently improving object detection and semantic segmentation. Moreover, improvements in both low-level reconstruction and high-level perception suggest that faithful recovery of infrared structures is closely related to reliable machine perception. These findings highlight the importance of preserving task-relevant thermal patterns and semantic structures in infrared super-resolution, providing a practical direction for developing restoration models that better serve downstream vision tasks.

\bibliography{faithir_references}

\clearpage
\newpage
\input{secs/appendix}

\end{document}

%% file: secs/appendix.tex
\maketitle
\appendix
\section*{Supplementary Contents}
This supplementary material follows the narrative order of the main paper. Detailed descriptions of each part are provided below.

\section{Training Details}
\label{app:training_details}
\noindent\textbf{Implementation Details.}
FaithIR adopts a patch size of 16. The Patch-Level Conditioning Branch contains 26 Patch-Level Conditioning Blocks (PCBs) with a hidden dimension of 1,152, while the Pixel-Level Restoration Branch contains four Pixel-Level Restoration Blocks (PRBs) with a pixel-token dimension of 16. During training, spatially aligned $256\times256$ HR patches and $64\times64$ LR patches are randomly cropped from each image pair, with the LR patches bicubically upsampled to the HR resolution as conditional inputs. FaithIR is trained from scratch on four NVIDIA A100 GPUs without pretrained natural-image generative models or image tokenizers. We optimize the conditional flow-matching objective using AdamW with a batch size of 32 and a cosine learning-rate schedule from $5\times10^{-5}$ to $5\times10^{-6}$. The model is trained for 500k iterations using BF16 mixed precision, gradient clipping with a maximum norm of 1.0, and an exponential moving average (EMA) with a decay rate of 0.9999.

% \noindent\textbf{Object Detection Settings on M3FD}
\subsection{Object Detection Settings on M3FD}

We employ YOLOv8n as the object detector and initialize it with the official pretrained weights. The detector is trained on 2,940 M3FD infrared images and evaluated on the 630 image validation set. All images are resized to $640\times640$. We train the detector for 50 epochs using AdamW with a batch size of 8 and an initial learning rate of $1\times10^{-3}$. After training, the detector parameters are fixed for evaluating the SR images generated by all compared methods.

% \noindent\textbf{Semantic Segmentation Settings on FMB}

\subsection{Semantic Segmentation Settings on FMB}

We employ SegFormer-B2, initialized with ImageNet-1K pretrained weights, for semantic segmentation. The model is trained on 1,220 labeled FMB images and evaluated on the 280-image test set. The input resolution is set to $512\times512$, and single-channel infrared images are replicated to three channels. We optimize the model using AdamW with an initial learning rate of $6\times10^{-5}$ and a weight decay of 0.01. The training objective combines cross-entropy and Dice losses with equal weights, and early stopping is employed during training. After training, the segmentation model is fixed for
evaluating all SR methods.

\section{Downstream Task Definitions on FLIR-IISR}
\label{FLIR-IISR-metrics}

Since FLIR-IISR does not provide ground-truth annotations for object detection or semantic segmentation, we employ fixed downstream models to evaluate the preservation of task-related information in the reconstructed images.

For detection, \textbf{Det. Num.} denotes the total number of detected objects over the 265 test images, while \textbf{Det. Rate} denotes the percentage of test images in which at least one object is detected.

For semantic evaluation, predictions on the corresponding HR images are used as semantic references. Let $p_c^{SR}$ and $p_c^{HR}$ denote the predicted pixel proportions of class $c$ in the SR and HR images, respectively. The segmentation deviation is defined as
\begin{equation}
\mathrm{Seg.\ Deviation}
=
\sum_c
\left|
p_c^{SR}-p_c^{HR}
\right|.
\end{equation}
A lower Seg. Deviation indicates that the semantic composition predicted from the SR image is more consistent with that obtained from its corresponding HR reference.

\section{SR Qualitative Results on M3FD and FMB}
\label{app:img_SR}
To further evaluate cross-dataset generalization, we provide additional qualitative comparisons on M3FD and FMB in this appendix. All models are trained on FLIR-IISR only and are directly tested on these unseen datasets without fine-tuning. As shown in the following figures, FaithIR produces more faithful thermal structures, clearer target contours, and more coherent infrared intensity distributions, while avoiding the over-smoothed results or artificial high-frequency details observed in some competing methods.
\begin{figure*}[t]
    \centering
    \includegraphics[width=\textwidth]{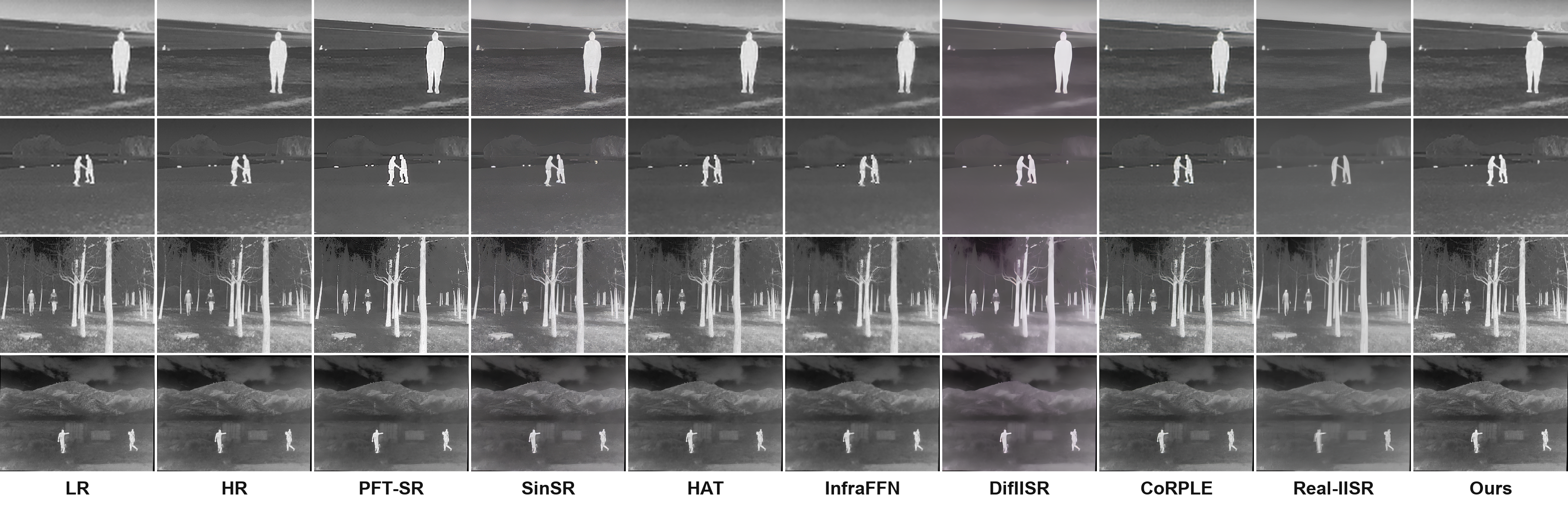}
    \caption{Qualitative comparison on the M3FD test set.}
    \label{fig:app_m3fd_sr}
\end{figure*}
\begin{figure*}[t]
    \centering
    \includegraphics[width=\textwidth]{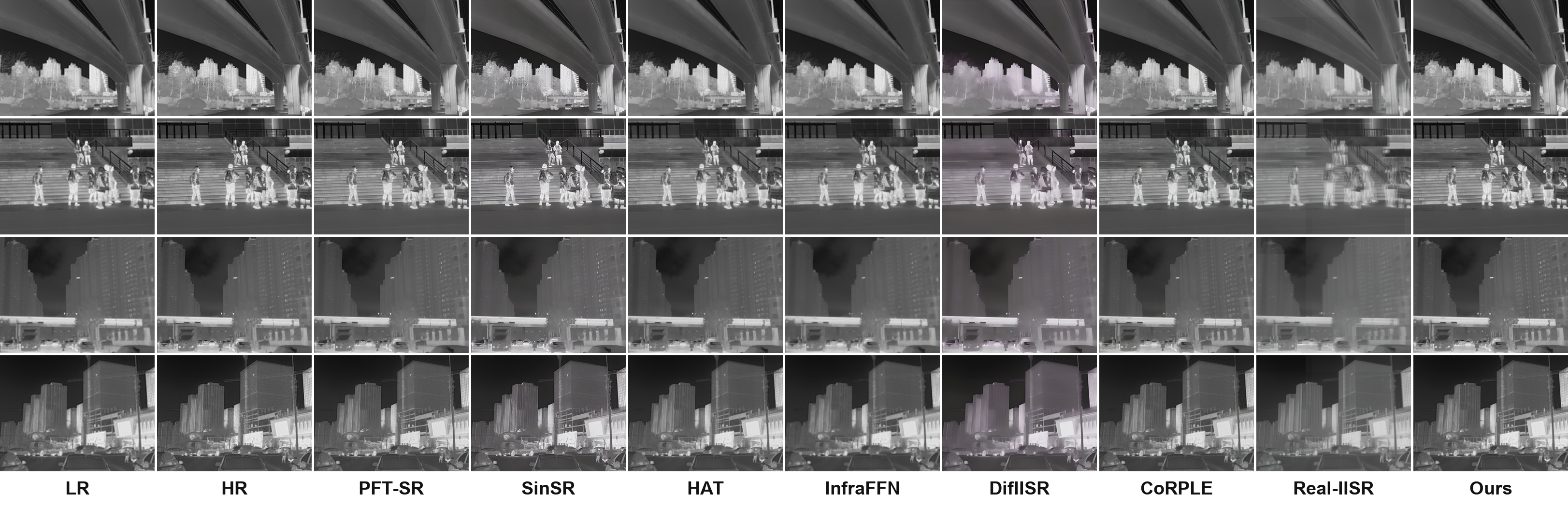}
    \caption{Qualitative comparison on the FMB test set.}
    \label{fig:app_fmb_sr}
\end{figure*}
As shown in Fig.~\ref{fig:app_m3fd_sr}, FaithIR better preserves human silhouettes, target boundaries, and local
thermal structures in diverse outdoor scenes. Compared with other methods, it avoids both excessive smoothing and unnatural high-frequency artifacts, leading to results that are more consistent with the HR infrared images.
As shown in Fig.~\ref{fig:app_fmb_sr}, FaithIR achieves more faithful recovery of building boundaries, pedestrian structures, and large-scale scene layouts. The reconstruction results maintain a smoother and more continuous infrared intensity distribution.

\section{Additional Object Detection Results}
\label{app:detection}

\begin{table*}[t]
\centering
\small
\setlength{\tabcolsep}{5.0pt}
\renewcommand{\arraystretch}{1.08}
\begin{tabular}{lcccccccc}
\toprule
Method 
& Det. Num. $\uparrow$
& $\mathrm{mAP}_{50}$ $\uparrow$
& People $\uparrow$
& Car $\uparrow$
& Bus $\uparrow$
& Motor. $\uparrow$
& Lamp $\uparrow$
& Truck $\uparrow$ \\
\midrule
InfraFFN
& 4,734
& 0.643
& 0.727
& 0.694
& 0.762
& 0.414
& 0.253
& 0.526 \\

DifIISR
& 4,338
& 0.602
& 0.660
& 0.686
& 0.743
& 0.357
& 0.124
& 0.504 \\

CoRPLE
& \underline{4,914}
& \underline{0.656}
& \underline{0.730}
& \underline{0.719}
& \textbf{0.802}
& \underline{0.471}
& \underline{0.291}
& \textbf{0.556} \\

Real-IISR
& 3,674
& 0.495
& 0.462
& 0.583
& 0.723
& 0.271
& 0.104
& 0.452 \\

\textbf{FaithIR (Ours)}
& \textbf{5,102}
& \textbf{0.661}
& \textbf{0.732}
& \textbf{0.724}
& \underline{0.782}
& \textbf{0.514}
& \textbf{0.324}
& \underline{0.533} \\
\bottomrule
\end{tabular}
\caption{Complete object detection results on M3FD.}
\label{tab:full_detection}
\end{table*}

\begin{figure*}[t]
    \centering
    \includegraphics[width=\textwidth]{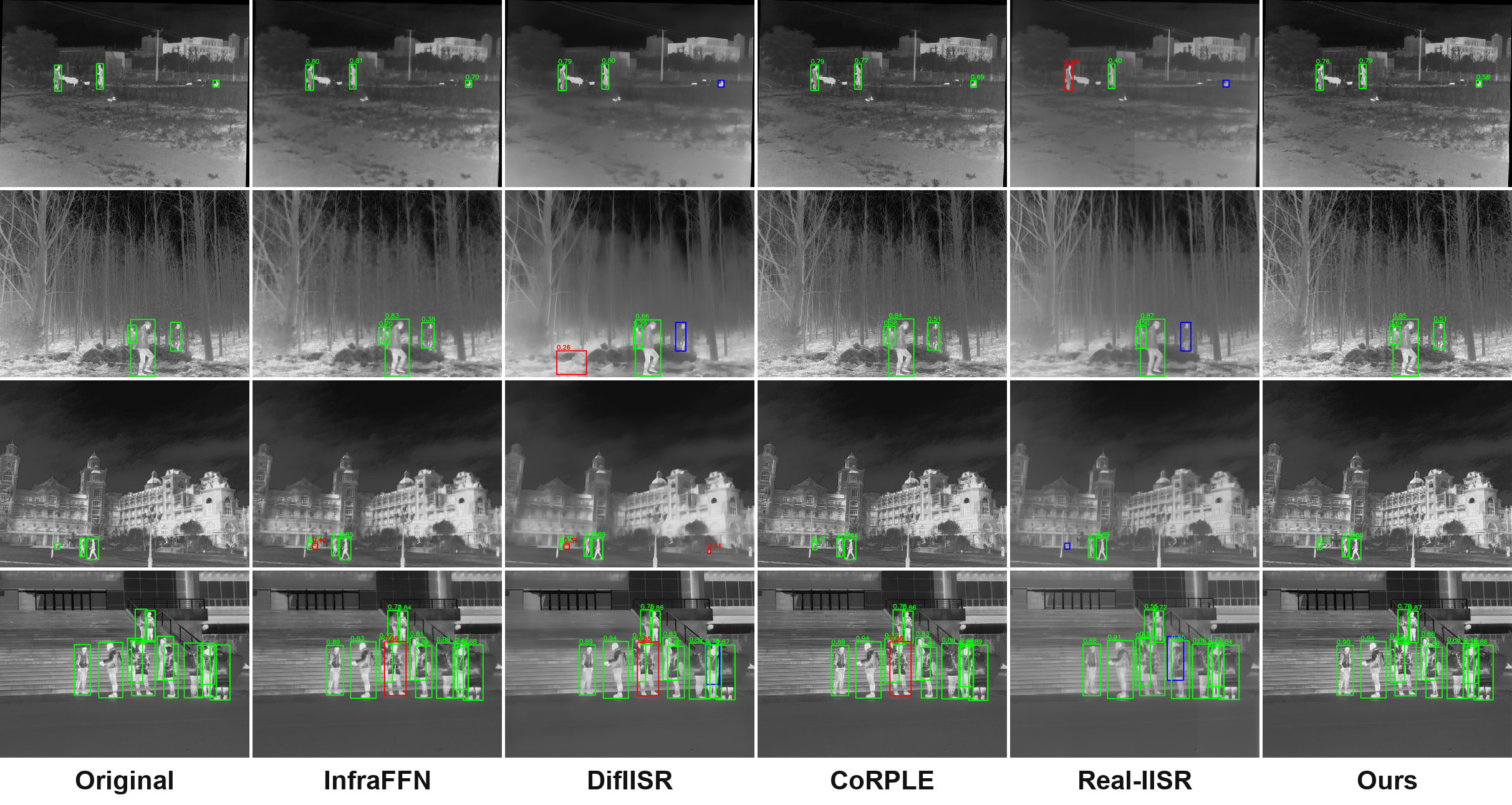}
    \caption{Additional object detection results on M3FD.}
    \label{fig:app_detection}
\end{figure*}
Table~\ref{tab:full_detection} reports the complete object
detection results on M3FD, including the total number of
detections, overall $\mathrm{mAP}_{50}$, and recall for all six
object categories. FaithIR achieves the highest number of detected objects and the best overall $\mathrm{mAP}_{50}$ among all SR methods. It also obtains the best recall for the People, Car, Motorcycle, and Lamp categories, while maintaining competitive performance for Bus and Truck. These results indicate that the improvement of FaithIR is not limited to a single target category, but is consistently reflected across small targets, vehicles, and roadside objects.
Figure~\ref{fig:app_detection} presents additional object detection comparisons on M3FD. The same fixed YOLOv8n detector is applied to the super-resolved images produced by all methods. FaithIR preserves small and low-contrast targets more reliably across diverse outdoor scenes, resulting in fewer false positives and missed detections. In particular, the restored pedestrian contours remain more complete in distant and cluttered regions, providing more reliable structural cues for the detector. In contrast, excessive smoothing or structural distortion in some competing methods may weaken target responses or introduce inaccurate detections.

\section{Additional Semantic Segmentation Results}
\label{app:segmentation}

\begin{table*}[t]
\centering
\small
\setlength{\tabcolsep}{3.8pt}
\renewcommand{\arraystretch}{1.05}
\resizebox{\textwidth}{!}{
\begin{tabular}{lccccccccccccccc}
\toprule
Method
& mIoU $\uparrow$
& PA $\uparrow$
& Road $\uparrow$
& Sidew. $\uparrow$
& Build. $\uparrow$
& Lamp $\uparrow$
& Sign $\uparrow$
& Veg. $\uparrow$
& Sky $\uparrow$
& Person $\uparrow$
& Car $\uparrow$
& Truck $\uparrow$
& Bus $\uparrow$
& Motor. $\uparrow$
& Pole $\uparrow$ \\
\midrule

InfraFFN
& 47.6
& 86.6
& 85.1
& 40.7
& 74.4
& 10.2
& 56.5
& 75.2
& \underline{88.0}
& \underline{58.3}
& \underline{74.8}
& 21.3
& 18.1
& 30.4
& \underline{17.3} \\

DifIISR
& 44.4
& 83.1
& 82.0
& 40.0
& 69.6
& 6.9
& 50.3
& 64.3
& 83.1
& 55.0
& 72.2
& 14.1
& \underline{21.8}
& 30.1
& 15.9 \\

CoRPLE
& \underline{48.6}
& \underline{86.9}
& \textbf{85.7}
& \underline{42.7}
& \underline{75.6}
& \underline{11.1}
& \textbf{57.9}
& \underline{75.7}
& 87.7
& \textbf{59.7}
& \textbf{75.1}
& 23.0
& 20.6
& \underline{32.5}
& 16.9 \\

Real-IISR
& 43.5
& 84.7
& 80.9
& 36.7
& 72.3
& 4.9
& 50.6
& 70.7
& 87.0
& 33.0
& 71.0
& \underline{24.0}
& \textbf{24.3}
& 22.6
& 15.0 \\

\textbf{FaithIR (Ours)}
& \textbf{49.0}
& \textbf{87.6}
& \underline{85.6}
& \textbf{43.2}
& \textbf{76.4}
& \textbf{11.6}
& \underline{56.6}
& \textbf{77.9}
& \textbf{89.2}
& 57.6
& 74.5
& \textbf{25.6}
& 19.5
& \textbf{33.1}
& \textbf{19.1} \\
\bottomrule
\end{tabular}}
\caption{Complete semantic segmentation results on FMB.}
\label{tab:full_segmentation}
\end{table*}

\begin{figure*}[t]
    \centering
    \includegraphics[width=\textwidth]{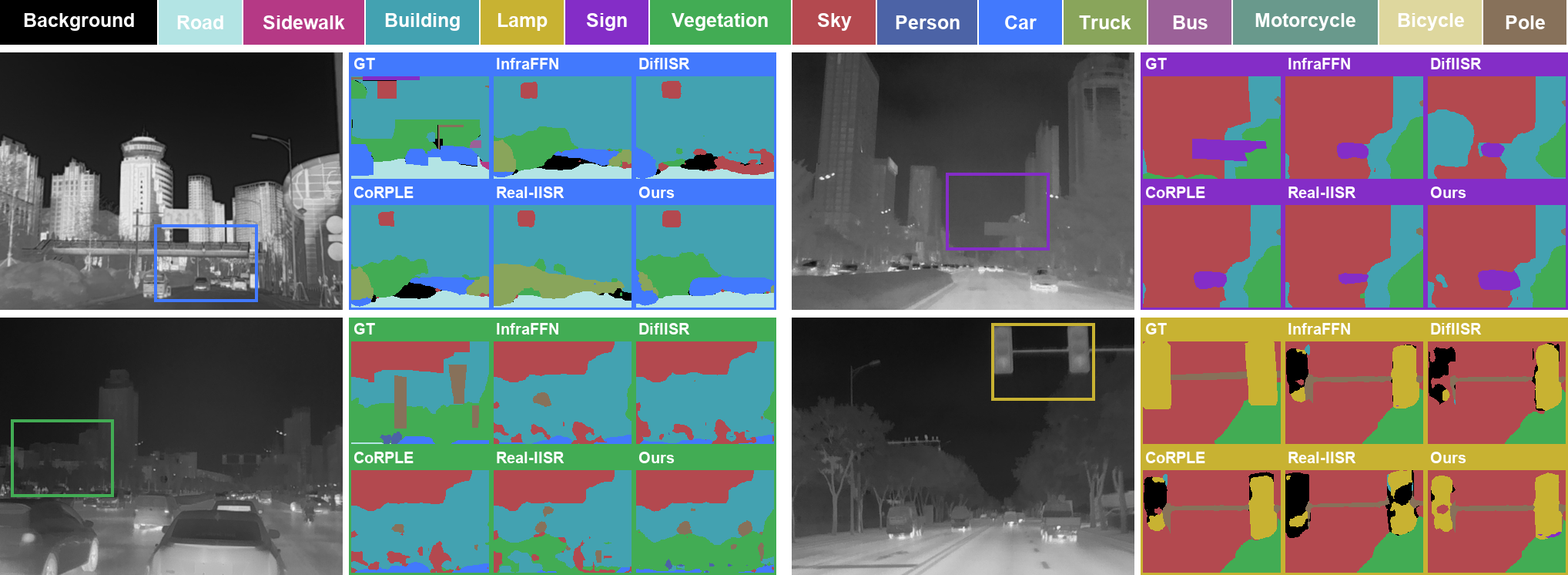}
    \caption{Additional semantic segmentation results on FMB.}
    \label{fig:app_segmentation}
\end{figure*}
Table~\ref{tab:full_segmentation} reports the complete semantic segmentation results on FMB, including mIoU, pixel accuracy (PA), and class-wise IoU for the reported semantic categories. FaithIR achieves the best overall mIoU and PA. It also obtains the highest IoU for Sidewalk,
Building, Lamp, Vegetation, Sky, Truck, Motorcycle, and Pole, while achieving the second-best performance on Road and Sign. These results demonstrate that FaithIR preserves both large-scale semantic regions and small object structures, leading to more reliable inputs for semantic segmentation. Figure~\ref{fig:app_segmentation} shows additional semantic segmentation comparisons on FMB using the same fixed SegFormer-B2 model. The enlarged regions cover challenging scene structures, including building boundaries, vegetation, roadside objects, and small semantic regions. FaithIR produces predictions that are more consistent with the ground-truth labels, with more complete regions and more accurate semantic boundaries. In comparison, competing methods may introduce fragmented predictions, boundary shifts, or incorrect class assignments, especially in low-contrast and spatially small regions. These results further demonstrate that faithfully restoring infrared structures benefits subsequent semantic understanding.

\section{Qualitative Results of the Ablation Study}
\label{app:ablation study}

To provide an intuitive view of the ablation results in Table~\ref{tab:ablation}, we present qualitative comparisons of the latent-space DiT baseline, the model without the Patch-Level Conditioning Branch, the patch-shared AdaLN variant, and the full FaithIR model.

\begin{figure*}[t]
    \centering
    \includegraphics[width=\textwidth]{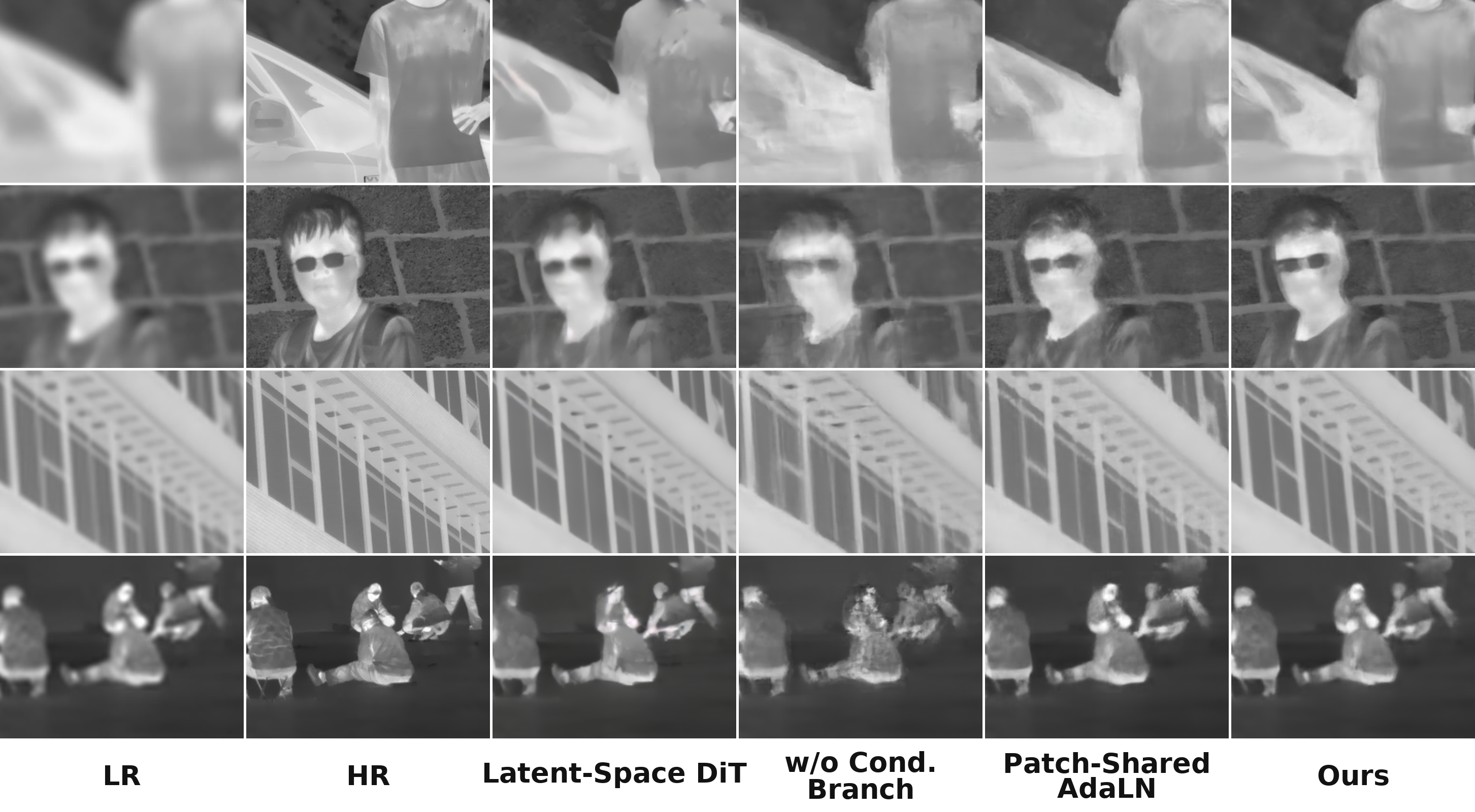}
    \caption{Qualitative SR results of different ablation variants on FLIR-IISR.}
    \label{fig:app_ablation_sr}
\end{figure*}

\begin{figure*}[t]
    \centering
    \includegraphics[width=\textwidth]{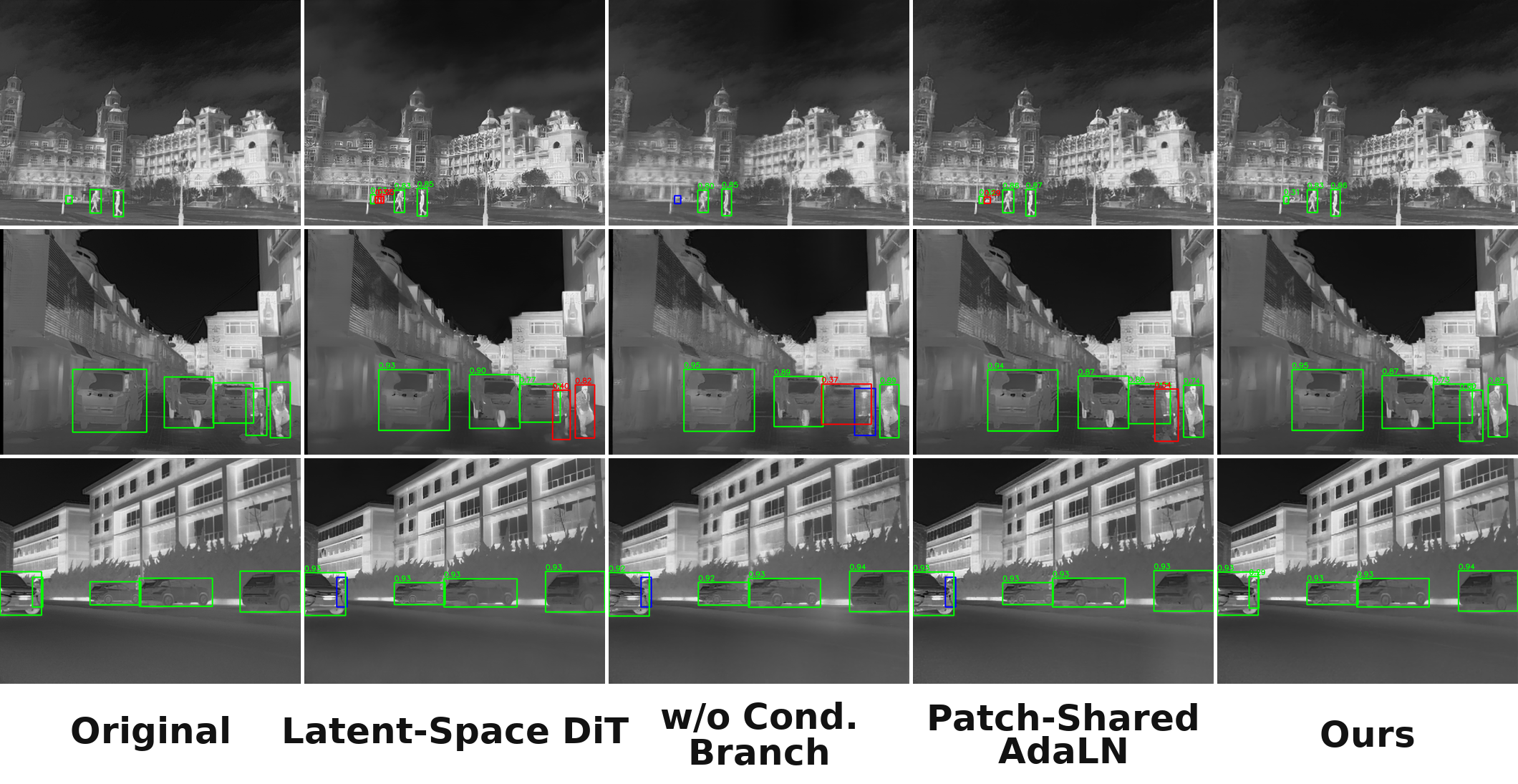}
    \caption{Qualitative object detection results of different ablation variants on M3FD.}
    \label{fig:app_ablation_det}
\end{figure*}

\begin{figure*}[t]
    \centering
    \includegraphics[width=\textwidth]{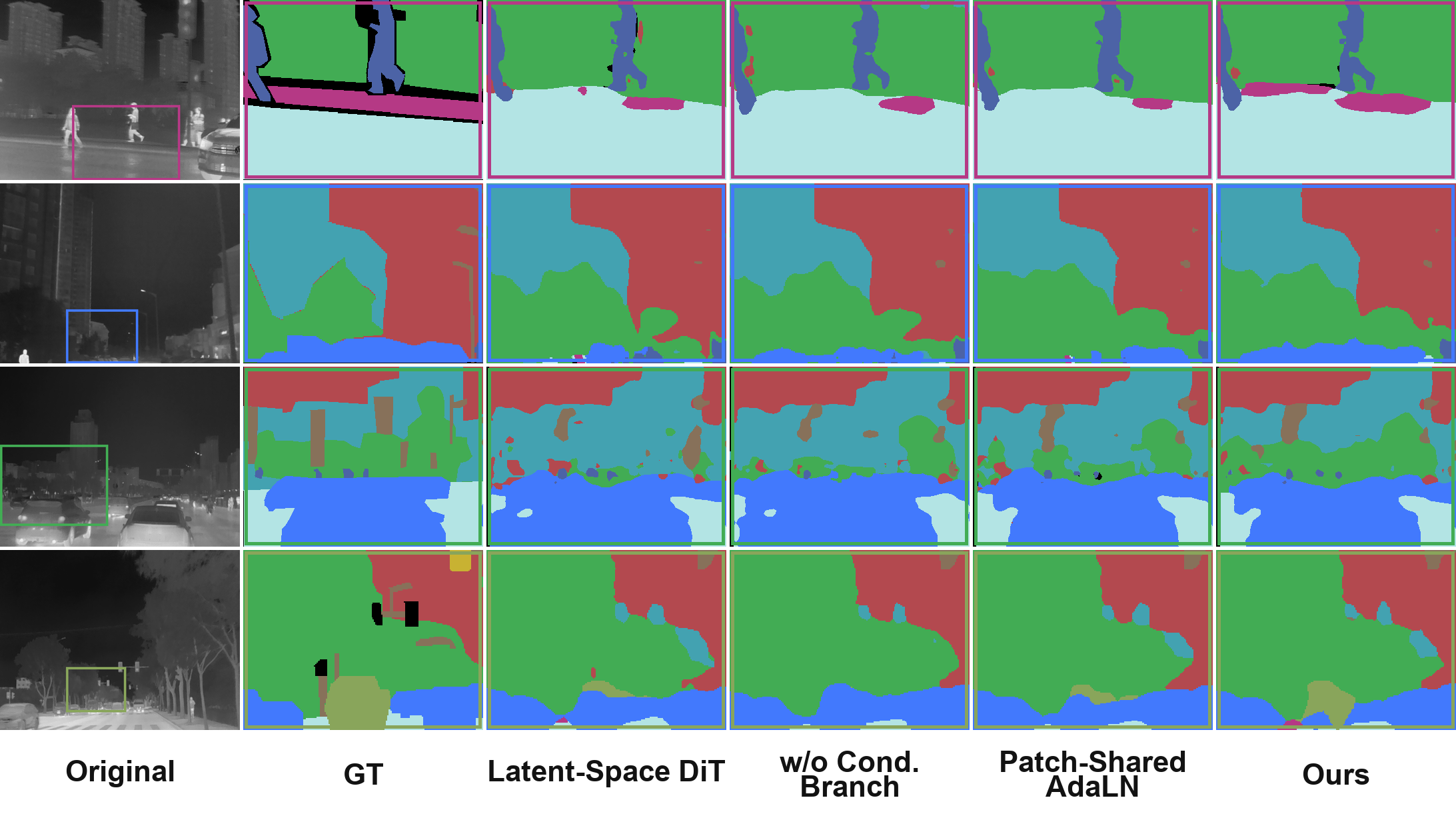}
    \caption{Qualitative semantic segmentation results of different ablation variants on FMB.}
    \label{fig:app_ablation_seg}
\end{figure*}

As shown in Fig.~\ref{fig:app_ablation_sr}, the latent-space baseline loses fine-grained thermal details, while removing the conditioning branch weakens structural guidance and patch-shared AdaLN limits adaptation to local variations. In contrast, FaithIR better preserves target contours and coherent thermal structures. Figures~\ref{fig:app_ablation_det} and~\ref{fig:app_ablation_seg} further show that FaithIR provides more reliable cues for object detection and produces more accurate semantic boundaries. These results are consistent with Table~\ref{tab:ablation} and demonstrate the joint benefits of pixel-domain restoration, patch-level conditioning, and pixel-wise modulation.
% To provide a more intuitive understanding of the ablation results in Table~\ref{tab:ablation}, we present qualitative comparisons of different FaithIR variants in terms of super-resolution reconstruction, object detection, and semantic segmentation. The compared variants include the
% latent-space DiT baseline, the model without the Patch-Level Conditioning Branch, the model using patch-shared AdaLN, and the full FaithIR model.

% As shown in Fig.~\ref{fig:app_ablation_sr}, the latent-space DiT baseline loses some fine-grained thermal details due to the additional encoding--decoding process. Removing the conditioning branch weakens the global structural guidance and produces less stable target contours, while patch-shared AdaLN provides insufficient adaptation to local variations within each patch. In contrast, the full FaithIR model better preserves human silhouettes, object boundaries, and coherent thermal distributions, producing results that are more consistent with the HR references.

% Figure~\ref{fig:app_ablation_det} further presents object detection results on M3FD using the same fixed YOLOv8n detector. Compared with the incomplete variants, the full FaithIR model provides more reliable structural cues for small and low-contrast targets, reducing inaccurate and missed detections. These visual results are consistent with the quantitative improvements reported in Table~\ref{tab:ablation}.

% As shown in Fig.~\ref{fig:app_ablation_seg}, the full FaithIR model also produces semantic segmentation predictions that are more consistent with the ground-truth labels. The latent-space baseline and incomplete variants exhibit less accurate boundaries, fragmented regions, or incorrect class assignments in some low-contrast areas. The results demonstrate that pixel-domain restoration, patch-level structural conditioning, and pixel-wise modulation jointly contribute to the preservation of task-relevant infrared structures.

